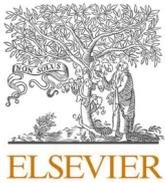
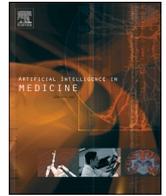

# Deep learning for autism detection using clinical notes: A comparison of transfer learning for a transparent and black-box approach

Gondy Leroy [a,*], Prakash Bisht [a], Sai Madhuri Kandula [a], Nell Maltman [b], Sydney Rice [c]

[a] *Management Information Systems, Eller College of Management, University of Arizona, Tucson, AZ, USA*
[b] *Speech Language and Hearing Sciences, College of Science, University of Arizona, Tucson, AZ, USA*
[c] *Pediatrics, College of Medicine, University of Arizona, Tucson, AZ, USA*

ABSTRACT

Autism spectrum disorder (ASD) is a complex neurodevelopmental condition whose rising prevalence places increasing demands on a lengthy diagnostic process. Machine learning (ML) has shown promise in automating ASD diagnosis, but most existing models operate as black boxes and are typically trained on a single dataset, limiting their generalizability.

In this study, we introduce a transparent and interpretable ML approach that leverages BioBERT, a state-of-the-art language model, to analyze unstructured clinical text. The model is trained to label descriptions of behaviors and map them to diagnostic criteria, which are then used to assign a final label (ASD or not). We evaluate transfer learning, the ability to transfer knowledge to new data, using two distinct real-world datasets. We trained on datasets sequentially and mixed together and compared the performance of the best models and their ability to transfer to new data. We also created a black-box approach and repeated this transfer process for comparison.

Our transparent model demonstrated robust performance, with the mixed-data training strategy yielding the best results (97 % sensitivity, 98 % specificity). Sequential training across datasets led to a slight drop in performance, highlighting the importance of training data order. The black-box model performed worse (90 % sensitivity, 96 % specificity) when trained sequentially or with mixed data.

Overall, our transparent approach outperformed the black-box approach. Mixing datasets during training resulted in slightly better performance and should be the preferred approach when practically possible. This work paves the way for more trustworthy, generalizable, and clinically actionable AI tools in neurodevelopmental diagnostics.

## 1. Introduction

Autism spectrum disorders (ASD) is a neurodevelopmental condition that is increasing in prevalence according to estimates from the Centers for Disease Control and Prevention (CDC)'s Autism and Developmental Disabilities Monitoring (ADDM) Network [1]. Treatments and interventions are the most impactful when they start early [2–6]. Unfortunately, the age of diagnosis is not decreasing and remains, on average, at four to five years old [1]. This may be partly due to the limited time that pediatricians have available to evaluate children. A 2023 review [7] showed that American pediatricians see an average of 69.5 patients per week. Furthermore, there is a shortage of specialists in developmental and behavioral pediatrics. In 2021, less than 70 % of developmental and behavioral pediatrics positions were filled, and in comparison to other specialties, the number of medical students choosing pediatrics is also declining [8]. Finally, for many children needing evaluation, the waitlist time is months long [9].

Machine learning (ML) can expedite the diagnosis of autism. However, existing ML projects to detect ASD share two shortcomings. The first is the use of black-box approaches, where the ML outcome is calculated using a computational approach that assigns a single label to a case without a clear rationale that is understandable to a human decision maker. It is difficult to assess the label's validity, especially for someone with limited clinical expertise, and it is even harder to adjust.

The second shortcoming is a lack of knowledge of the applicability of the algorithms to new data. A model trained on one dataset will show a different performance when transferred to another dataset because datasets may differ in the language used (e.g., expressions used in rural clinics versus university hospitals), the focus of children's evaluations (e.g., speech versus behaviors), or the composition of the cohort (e.g., age and gender), among other factors. The challenge of these transfers is rarely addressed.

We developed a transparent approach to diagnose cases of ASD. We focus on constituent steps in the diagnostic process, i.e., identifying descriptions of key behaviors related to ASD and labeling these using the Diagnostic and Statistical Manual of Mental Disorders (DSM-5) [10]. A case decision is made using the DSM-5 guidelines, requiring deficits in the three areas of social communication and interaction, and at least two types of restricted, repetitive behaviors [11]. In earlier work, we created an ensemble of ML algorithms [12] which we have replaced here with one BioBERT model. We aim to provide two contributions. First, we evaluate the model's transferability using different datasets. With the increasing number of diverse settings, such as different hospitals, clinics, and EHR systems, this is an important issue. Second, we compare our transparent approach with a black-box version. With long wait times and limited developmental professionals, a transparent approach can contribute to supporting early diagnosis of ASD and be especially useful for those who are not specialists.

## 2. Related work

### 2.1. Machine Learning for mental health

ML is used in all aspects of healthcare and medicine, with numerous peer-reviewed publications specifically focused on the subdomain of neurodevelopment. These ML projects use a variety of structured data, process unstructured data to create structure, and increasingly, with the availability of deep learning models, use the unstructured data itself. A 2021 review of the role of natural language processing (NLP) in ML showed that it was often used to extract features or to classify the entire case, to process the patient's language, or to provide insights into conditions [13]. With big data, ML has primarily focused on diagnosis, treatment, public health, research, and administration [14], but it has also been leveraged to extract information from social media data. For example, several ML projects focus on Twitter data to label people. In this case, it often requires a medical-oriented survey to establish the ground truth to enable the ML training [14].

In recent years, deep learning algorithms, such as convolutional neural networks (CNN), long short-term memory (LSTM) networks, and transformers, have become available through open-source repositories, e.g., HuggingFace[1] or Data Bricks,[2] and are readily used. These advanced algorithms, combined with relatively cheap storage and compute time, enable advanced ML on complex datasets such as imaging and genetic data [15] which are now also increasingly used in neurodevelopmental conditions. Moreover, with the availability of large language models (LLMs), a multitude of new approaches with chatbots, avatars, and animal robots [16] are rapidly being developed and tested.

However, ML still suffers from several shortcomings and is limited by a variety of obstacles, e.g., the need for appropriate data to tune models to a specific context [17], avoiding the creation of misinformation [18], and the need for more fairness by avoiding algorithm bias, e.g., gender bias [19]. ML models can be categorized as black-box (the rationale for a decision is not or cannot be given or understood) and white-box approaches (a domain expert can understand the rationale). For important and impactful decisions, black-box approaches are suboptimal, and an increasing number of projects focus on developing explainable, interpretable, or transparent AI and ML. We use the term "transparent" with the understanding that the definitions are still in flux. We believe this term describes our process best since our final ASD label can be easily deduced from the diagnostic criteria used to label the child's behaviors.

Specifically for ASD, most projects show the strengths and weaknesses described above. They assign a single label, ASD or no ASD (some assign a more fine-grained label), and focus on comparing algorithms [20]. The data is extracted from screening instruments, surveys, or clinician-provided data [21–23]. Sometimes, data is algorithmically extracted from text [24], but often, the unstructured data is manually transformed, e.g., by expert tagging of behaviors in home videos [21]. In the last few years, with deep learning, new data sources are increasingly leveraged. One project focused on early prediction and found that maternal emotion and nutrition were important risk factors [25]. Others focus on regions in the brain, e.g., using MRI to associate gray matter volume with autism severity [26] or detecting abnormalities in ASD versus control children [27]. Different workflows are being designed and tested for brain image-based diagnosis to avoid undue influence by outliers and to avoid bias [28]. Data is also increasingly combined, e.g., through advanced convolutional networks (CNN) using different types of imaging [29]. Genomic data is also used, e.g., proteome and metabolome biomarkers for ASD [30], as well as continuous data such as electroencephalography (EEG) signals [31] or eye-movement data [32,33]. In our own projects, we are experimenting with synthetic data sets [34].

### 2.2. Transfer Learning

The term transfer learning (TL) is commonly used to describe the transfer of ML models that have been trained for one task, typically after tuning or making other adjustments, and are then applied to a new task. The underlying model or some of its layers are tuned for the different outcomes or tasks. Most TL is homogeneous and the application domain remains the same, e.g., image analysis. The concept originates in cognitive research, and TL techniques were surveyed as early as 2010 [35]. Especially with deep learning algorithms and large language models, which require large amounts of data and processing time, the ability to reuse and adjust an existing model is very valuable. Not surprisingly, this is a well-studied topic with more than 45,000 publications in PubMed in one year (2024).

In medicine, several reviews of TL have been completed. TL for image classification with CNNs [36] showed that projects mainly focused on testing different algorithm combinations. The main finding was that starting with feature extraction was better than fine-tuning, which is also more computationally expensive. Another review of transferring CNNs from non-medical to medical images [37] concluded that performance was overall reasonable. When datasets were smaller (<1000 images), feature extraction was used more often, while with larger datasets, fine-tuning was used more often. A review of TL for EEG in brain-computer interface applications [38] It was found that tasks were usually limited to classification and regression. While deep learning was very promising as a starting point for TL, the datasets were often small, and the computational burden was too high for practical applications. Another common focus of TL is molecular data since these data are large and complex, and there is a benefit from being able to reuse models. TL was shown to have high potential when working with molecular data across models (e.g., human versus mouse) [39].

Fewer studies have been published in the area of neurological conditions. One exception is a study that demonstrated the benefits of TL for neuroimaging datasets related to Alzheimer's, offering several advantages over training from scratch. However, there is a real potential for overfitting with small datasets relative to the number of parameters that can be tuned [40]. Specifically for autism, TL from 2D to 3D brain images improved ASD classification [41]. TL using feature extraction and CNNs showed excellent results in detecting ASD in electroencephalography signals, even though it was a small dataset of 45 cases [42]. However, others point out the difficulty of transferring ML, e. g., difficulties when training on emotions in images from adults and transferring to images of children to recognize ASD [43].

---

[1] https://huggingface.co/
[2] https://www.databricks.com/

## 3. Methods

### 3.1. Data sets

We utilize clinical notes from two datasets that contain information on children with neurodevelopmental conditions. Table 1 shows an overview of the data and the number of behavioral descriptions available for each diagnostic criterion (a detailed description of each criterion is provided in Appendix A).

The first dataset was gathered by the Centers for Disease Control and Prevention (CDC) Autism and Developmental Disabilities Monitoring (ADDM) Network and contains their records for 8- and 4-year-old children living in Maricopa County, Arizona, in 2016. All cases involved children who underwent a clinical or educational evaluation due to concerns about autism. The original surveillance dataset was labeled using DSM-IV-R criteria for diagnosis. However, because diagnostic criteria had been updated since then, a subset of 200 cases was relabeled using DSM-5 criteria. Line-level re-labeling with individual criteria was conducted by the CDC labeler, who was part of the original CDC surveillance. The line and case labels assigned by the CDC-labeler and associated with the DSM-5 criteria are used to evaluate our algorithms. Our approach is trained on individual sentences, comprising 44,429 sentences in this dataset, with 10.9 % receiving a criterion label, and 68 % of cases are labeled as ASD.

The second dataset was extracted from the University of Arizona Clinical Data Warehouse (CDW) and comprises EHR with ICD10 diagnostic codes related to autism (F84.0 Childhood Autism, F84.5 Asperger Syndrome, F84.1 Atypical Autism, F84.8 Other pervasive developmental disorders, F84.9 Pervasive developmental disorders, unspecified, F84.4 Overactive disorder associated with mental retardation and stereotyped movements, F84.2 Rett's Syndrome, F84.3 Childhood Disintegrative Disorder) and one ICD10 screening code (Z13.41 – encounter for autism screening). The cases with an F84.0 label (19.83 % of our cases) were considered ASD to evaluate our algorithms. A team of trained master's-level clinicians with ASD expertise labeled the individual lines in the dataset. Coders were trained to the reliability of 80 % or greater on case criteria (i.e., ASD vs. no ASD) and individual DSM-5 criteria (A1–3, B1–4) at 90 % or greater according to the same CDC standards as for the ADDM records. There were 101,174 lines in this dataset, of which 4.6 % received an ASD criterion label.

### 3.2. Transfer Learning design

Fig. 1 shows our use of 10-fold cross-validation. We use a simple approach where the model is trained on N-1 folds and tested on the remaining test fold. This is repeated with each fold serving once as the test fold. After performing this process with all folds, we calculate the average performance. To evaluate the impact of new data and to have static comparison points, we select the best model from the 10 training rounds and its associated test fold. The best models from each dataset are used for further tuning and the associated test folds serve as comparison points.

**Table 1**
Case and criterion-level counts of data in the two datasets (ADDM: Centers for Disease Control and Prevention (CDC) Autism and Developmental Disabilities Monitoring (ADDM) Network, CDW: Clinical Data Warehouse).

|  | ADDM | CDW EHR |
|---|---|---|
| Line Count | 44,429 | 101,174 |
| Case Count | 200 | 600 |
| Case ASD (%) | 68 | 19.83 |
| Criteria Count |  |  |
| A1 | 1167 | 969 |
| A2 | 611 | 712 |
| A3 | 738 | 698 |
| B1 | 692 | 830 |
| B2 | 485 | 410 |
| B3 | 226 | 250 |
| B4 | 917 | 809 |
| Total | 4836 | 4678 |

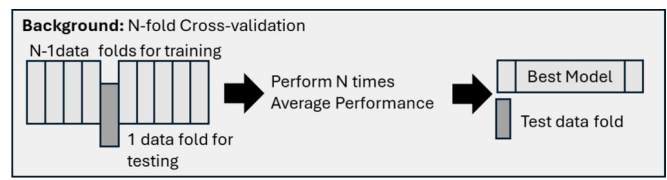

**Fig. 1.** Model training using N-fold cross validation: average performance and best model.

As shown in Fig. 2, we apply 10-fold cross-validation to the three datasets: ADDM, CDW EHR, and the combined dataset. We first report the average performance and then set aside the best model for each dataset. For example, the best model for the ADDM dataset was found when training on folds 1–3 and 5–10 with test fold 4, for the CDW EHR dataset, the best model was found with test fold 9, and for the mixed dataset with test fold 7. These best models are then tested on all three test folds.

As shown in Fig. 3, the best models are further tuned using a second, different dataset. For the mixed dataset, no additional training is conducted. After this second round of tuning using 10-fold cross-validation, we report average performance on the new dataset. Again, we select the best model and apply it to the original test folds from the first round. For example, the best tuned ADDM-EHR model (after training on both datasets) is applied to the original test folds set aside after initial training, allowing us to compare performance after training on one, two, or mixed datasets.

For our transparent approach, we evaluate performance at the criterion level, i.e., how well individual sentences were labeled with A1–3 and B1–4 DSM criteria. Evaluation metrics for this criterion-level data include precision, recall, and the score on the F1 measure. Evaluation metrics at the case level include accuracy, sensitivity, and specificity.

For our black-box approach, we report only accuracy, sensitivity, and specificity at the case level, since no criterion level data are available.

### 3.3. Transparent and black-box approach comparison

We conduct the process described above using our transparent and black-box approach.

Our transparent approach focuses on identifying individual behaviors described in a record that can be labeled with a relevant DSM criterion, as well as a case-level label. We algorithmically combine the information at the case level and assign an ASD label when a case contains examples of the three A criteria (deficits in social communication and interaction) and at least two of the B criteria (restricted and repetitive behaviors) [11].

For the black-box model, the text is split into sentences, and the first 512 tokens per sentence are used as input. Most ML approaches use limited free text input, e.g., the first 512 tokens of an entire text, and some compare different selection schemes, such as the first section, the last section, or a combination of these, as input [44]. However, limiting the input to 512 tokens per case would not provide a reasonable comparison for our two approaches. We therefore use all sentences as input. The final labeling with ASD is assigned by calculating the percentage of lines with an ASD label for a case. We compare different thresholds to assign the label: 0.2, 0.4, 0.5, 0.6, 0.8. A threshold of 0.2 means that when an ASD label is assigned to 20 % of sentences, the case itself is labeled as ASD. This is a lenient approach to assigning the ASD label. A threshold of 0.8 means that when an ASD label is assigned to 80 % of sentences, the case is labeled as AD, which is a much higher barrier and stricter approach.

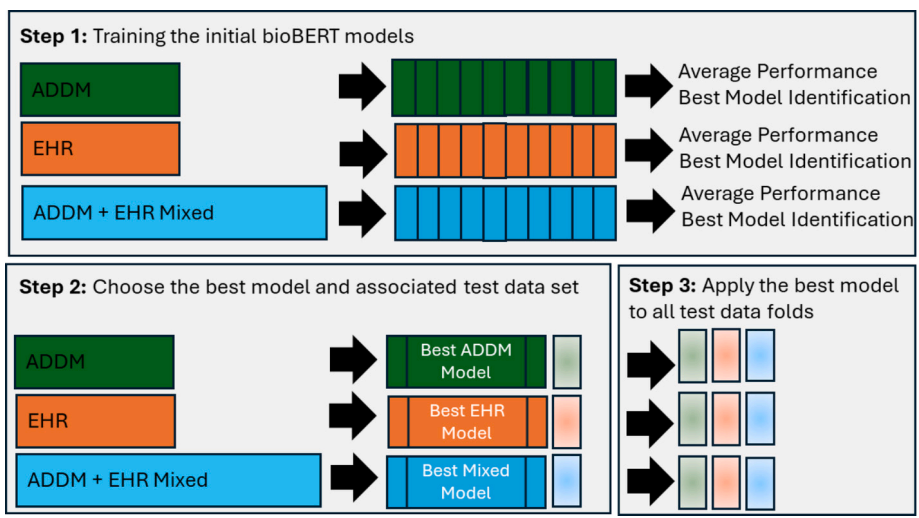

**Fig. 2.** Model training for each dataset using N-fold cross-validation.

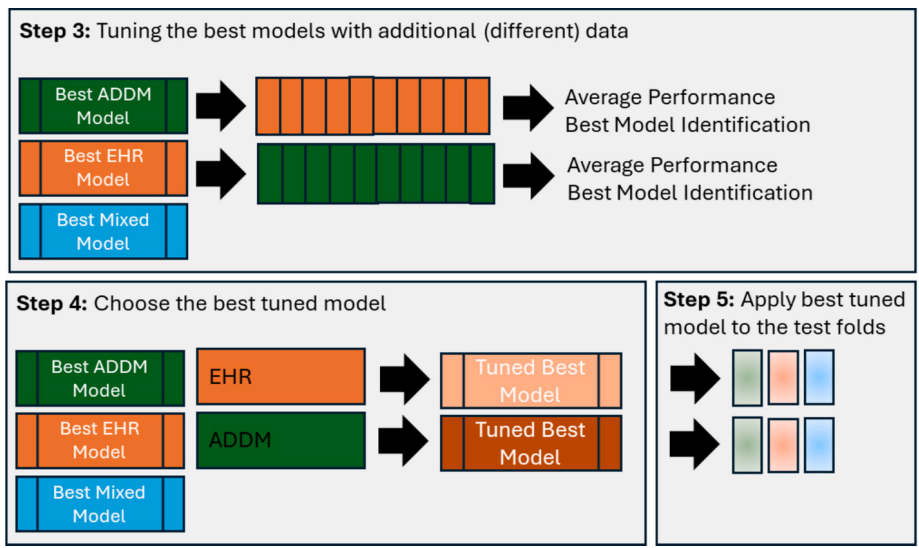

**Fig. 3.** Model tuning using additional (different) datasets.

### 3.4. BioBERT model

In previous work, we combined one rule-based and three machine-learning models in a variety of ensembles [45]. Here, we have replaced these algorithms with the BioBERT (version 1.2) model. Both the transparent and black-box versions use the same parameters, facilitating comparison. The model parameter details are listed in Table 2. The learning rate differed from the tuning rate so that the first trained model would not get overpowered by subsequent learning. This helps avoid the erasure of previous knowledge. By allowing a high number of epochs, we believe enough learning will ensue during tuning.

The transparent approach was implemented on a Paperspace environment equipped with 45GB RAM, 8 CPUs, and a 24GB GPU. Each epoch required, on average, approximately 1 h, resulting in a total training time of 70 to 80 h per model.

The black-box models were trained on the University of Arizona's High Performance Computing (HPC) system, utilizing 5 GB of RAM, 1 CPU, and four V100 GPUs (each with 32 GB of RAM). The training duration per epoch remained consistent, averaging approximately 1 h to 1 h and 15 min, with total training time ranging between 70 and 90 h per model.

**Table 2**
BioBERT model parameters.

| | |
|---|---|
| Libraries | Python 3.9.16. libraries: beautifulsoup4 (v4.11.1), chardet (v3.0.4), contractions (v0.1.73), huggingface-hub (v0.12.0), keras (v2.9.0), Keras-Preprocessing (v1.1.2), matplotlib (v3.6.1), matplotlib-inline (v0.1.6), openpyxl (v3.0.3), pandas (v1.5.0), numpy (v1.23.4), regex (v2022.10.31), scikit-learn (v1.1.2), seaborn (v0.12.0), tensorflow (v2.9.2), transformers (v4.21.3), tqdm (v4.64.1), and unicodedata (v13.0.0) |
| Tokenizer | BioBERT Tokenizer |
| Learning rate | $1 \times 10^{-5}$ |
| Turning rate | $5 \times 10^{-7}$ |
| Activation Function | Sigmoid |
| Epochs | 8 for each fold |
| Stopping | Early stopping by monitoring validation loss. If the validation loss doesn't improve by a minimum value min_delta = 0.007 for a consecutive number of 4 epochs (i.e., the 'patience'), then training is stopped. |
| Min_delta | 0.007 |
| Patience | 4 |
| Loss function | Binary cross-entropy |
| Optimizer | The default included in the library: ADAM with the number of warmup steps at 20 % of the training steps |

## 4. Results

### 4.1. Transfer Learning for the transparent approach

#### 4.1.1. Criterion results

Table 3 shows the overall results after N-fold cross-validation training of sequential and mixed data. Following the first round of training, we observe comparable numbers for precision and recall across all datasets. The average precision is 69 %, 67 %, and 69 % for the ADDM, EHR, and mixed datasets, respectively; recall is slightly lower, at 60 %, 55 %, and 54 %. The F1 measure displays a similar pattern, with scores of 0.64, 0.60, and 0.61, respectively.

The best performance per diagnostic criterion differs slightly in the datasets, but only by a few percentage points. For the ADDM dataset, precision (78 %) and recall (76 %) are highest for A2. For the EHR dataset, precision is highest for B3 (74 %), and recall is highest for B4 (68 %). For the mixed dataset, precision is highest for B2 (74 %), and recall is highest for A2 (71 %). Criterion B3 is poorly recognized in each dataset, with low recall rates of 39 % in the ADDM dataset, 27 % in the EHR dataset, and 29 % in the mixed dataset.

When further tuning is applied to the best models using the second dataset, the overall performance decreases slightly for both Sequence 1 (where training is first performed on ADDM data and then on EHR data) and Sequence 2 (where training is first performed on EHR data and then on ADDM data). Overall, in Sequence 1, precision decreases from 69 % to 63 %, recall decreases from 60 % to 54 %, and the score for the F1 measure decreases from 0.64 to 0.58. In Sequence 2, precision increases slightly from 67 % to 68 %; however, recall decreases from 55 % to 48 %, and consequently, the score for the F1 measure decreases from 0.60 to 0.54.

The mixed data model performs slightly better than the sequentially trained models with 64 % precision, 54 % recall, and a 0.61 score on the F1 measure. Since our goal is to evaluate transfer learning, we compared the mixed data model with the final sequentially trained models from Sequence 1 and Sequence 2 for precision, recall, and the F1-measure using paired-sample, two-tailed *t*-tests. Among the six t-test, there were two showing significant differences: the precision of the mixed model (69 %) is significantly better than that of the final Sequence 1 model (63 %), t(6) = 2.77, p = 0.032, and the recall of the mixed model (54 %) is significantly better than that of the final Sequence 2 model (48 %), t(6) = 4.86, p = 0.0028. If we apply a Bonferroni correction for six tests, the required *p*-value would be lowered to 0.008, and only the second effect would be considered significant.

Table 4 shows results for the best individual models. For example, for the ADDM dataset, the best model was trained on folds 1–3 and 5–10 and tested on fold 4, for the EHR, it was the model associated with test fold 9; and for the mixed data, it was the model associated with test fold 7.

The models show good performance on their own data. The ADDM model (with test fold 4) reaches 71 % precision, 67 % recall, and scores 0.69 on the F1 measure. The best model on EHR data (with test fold 9) reaches 70 % precision, 70 % recall, and scores 0.70 on the F1 measure. The best model for the mixed data (with test fold 7) reaches 73 % precision, 59 % recall, and scores 0.65 on the F1 measure.

When these best models are applied to the comparison data, performance is lower except for the mixed data model. For example, the best EHR model performs very poorly, with 40 % precision, 34 % recall, and a score of 0.37 on the F1 measure on the ADDM test data. In contrast, the best mixed model performs much better, with 74 % precision and 63 % recall, and scores 0.68 on the F1 measure on this ADDM test data. A similar pattern emerges when comparing performance on the EHR test data: the best ADDM model achieves only 49 % precision and 45 % recall and scores 0.47 on the F1 measure, while the mixed data model performs better on this data with 63 % precision, 50 % recall, and a score of 0.56 on the F1 measure.

With further tuning, the individual models exhibit different behaviors. Sequence 1 demonstrates that the best ADDM model's performance deteriorates with the inclusion of additional EHR (different) data: precision decreases from 71 % to 59 %, recall from 67 % to 66 %, and the F1 score decreases from 0.69 to 0.62. In contrast, Sequence 2 shows that the best EHR model's performance increases when additional ADDM (different) data is used: precision increases from 70 % to 98 %, recall increases from 70 % to 83 %, and the F1 score increases from 0.70 to 0.90.

These results demonstrate that focusing on one fold yields a different overall impression than using N-fold averages. The order of the datasets leads to differences in performance. Sequentially trained models do not uniformly improve with more similar or different data.

#### 4.1.2. Case results

Table 5 shows the case-level results. For the macro-average (N-fold average), we see that the mixed data model achieves the best performance overall with 97 % sensitivity and 98 % specificity. The models in sequence 1 and 2 perform initially well, but there is deterioration in some measures after adding the second dataset. Accuracy (98 %) is the same for all three datasets. With further tuning, the performance generally decreases on all measures. For example, in sequence 1, sensitivity decreases from 99 % to 93 %, and accuracy from 98 % to 97 %. Sensitivity increases slightly in Sequence 1 from 96 % to 97 %. In sequence 2, further tuning on ADDM data for the EHR model leads to lower scores on all metrics.

**Table 3**
Transparent model results at criterion level for sequential and mixed data training.

|  | Sequence 1 | | | Sequence 2 | | | | | |
|---|---|---|---|---|---|---|---|---|---|
| Initial Training on | **ADDM** | | | **EHR** | | | **Mixed** | | |
| Label | **Precision** | **Recall** | **F1** | **Precision** | **Recall** | **F1** | **Precision** | **Recall** | **F1** |
| A1 | 0.63 | 0.53 | 0.58 | 0.56 | 0.60 | 0.58 | 0.68 | 0.55 | 0.61 |
| A2 | 0.78 | 0.76 | 0.77 | 0.66 | 0.64 | 0.65 | 0.70 | 0.71 | 0.70 |
| A3 | 0.68 | 0.57 | 0.62 | 0.65 | 0.57 | 0.61 | 0.69 | 0.43 | 0.53 |
| B1 | 0.68 | 0.65 | 0.66 | 0.71 | 0.58 | 0.64 | 0.70 | 0.66 | 0.68 |
| B2 | 0.69 | 0.64 | 0.66 | 0.67 | 0.51 | 0.58 | 0.74 | 0.53 | 0.62 |
| B3 | 0.72 | 0.39 | 0.51 | 0.74 | 0.27 | 0.40 | 0.67 | 0.29 | 0.40 |
| B4 | 0.65 | 0.62 | 0.63 | 0.68 | 0.68 | 0.68 | 0.66 | 0.60 | 0.63 |
| Macro Average | **0.69** | **0.60** | **0.64** | **0.67** | **0.55** | **0.60** | **0.69** | **0.54** | **0.61** |
| Further Tuning on | **EHR** | | | **ADDM** | | | | | |
| A1 | 0.54 | 0.57 | 0.55 | 0.66 | 0.45 | 0.54 | | | |
| A2 | 0.65 | 0.70 | 0.67 | 0.85 | 0.65 | 0.74 | | | |
| A3 | 0.60 | 0.50 | 0.55 | 0.66 | 0.43 | 0.52 | | | |
| B1 | 0.70 | 0.61 | 0.65 | 0.70 | 0.58 | 0.63 | | | |
| B2 | 0.64 | 0.54 | 0.59 | 0.70 | 0.47 | 0.56 | | | |
| B3 | 0.58 | 0.28 | 0.38 | 0.53 | 0.25 | 0.34 | | | |
| B4 | 0.69 | 0.60 | 0.64 | 0.63 | 0.52 | 0.57 | | | |
| Macro Average | **0.63** | **0.54** | **0.58** | **0.68** | **0.48** | **0.56** | | | |

**Table 4**
Transparent model results for best models for sequential and mixed data training.

|  | Sequence 1 | | | Sequence 2 | | | | | |
| --- | --- | --- | --- | --- | --- | --- | --- | --- | --- |
| Initial Training on | **ADDM** | | | **EHR** | | | **Mixed** | | |
| Test fold used for comparing models | **ADDM fold 4** | | | **EHR fold 9** | | | **Mixed Fold 7** | | |
|  | Precision | Recall | F1 | Precision | Recall | F1 | Precision | Recall | F1 |
| Best ADDM Model | 0.71 | 0.67 | 0.69 | 0.49 | 0.45 | 0.47 | 0.75 | 0.63 | 0.68 |
| Best EHR Model | 0.40 | 0.34 | 0.37 | 0.70 | 0.70 | 0.70 | 0.70 | 0.46 | 0.56 |
| Best Mixed Model | 0.74 | 0.63 | 0.68 | 0.63 | 0.50 | 0.56 | 0.73 | 0.59 | 0.65 |
| Further Tuning on | **EHR** | | | **ADDM** | | | | | |
| Best ADDM Model | 0.59 | 0.66 | 0.62 | 0.76 | 0.49 | 0.60 | | | |
| Best EHR Model | 0.83 | 0.88 | 0.85 | 0.98 | 0.83 | 0.90 | | | |

**Table 5**
Transparent model results at case level for sequential and mixed data training.

|  | Sequence 1 | | | Sequence 2 | | | | | |
| --- | --- | --- | --- | --- | --- | --- | --- | --- | --- |
| Initial Training on | **ADDM** | | | **EHR** | | | **Mixed** | | |
|  | Accuracy | Sensitivity | Specificity | Accuracy | Sensitivity | Specificity | Accuracy | Sensitivity | Specificity |
| N-fold Average | 0.98 | 0.99 | 0.96 | 0.98 | 0.98 | 0.98 | 0.98 | 0.97 | 0.98 |
| Further Tuning on | **EHR** | | | **ADDM** | | | | | |
| N-fold Average | 0.97 | 0.93 | 0.97 | 0.86 | 0.82 | 0.91 | | | |

Table 6 shows the comparison of the best folds. We see consistently high values after training on the mixed dataset. Notably, in sequence 2, where training was first performed on EHR and then tuned on ADDM, even the best fold's performance (i.e., the ADDM best fold) achieves only 69 % sensitivity and 86 % specificity. The best model from the mixed dataset also achieves better accuracy in sequence 1 but not sequence 2 (96 % versus 97 % for the mixed and EHR model respectively).

*4.2. Transfer Learning for the black-box approach*

Table 7 shows the results for the black-box approach. Since we only assign an ASD label, no criterion-level results can be shown. Overall, all performance measures are lower for the black-box model than for the criterion-based, transparent model after training on the first dataset, regardless of the dataset used. As expected, there is a progression of lower sensitivity with a higher threshold, i.e., if more lines need to be labeled as ASD, then fewer records are labeled as such. Specificity increases with a higher threshold, but is especially low with the ADDM dataset.

When training the best ADDM model from sequence 1 on EHR data, the performance increases enormously, especially with lower thresholds. This is not the case for sequence 2. After training on EHR data, further training of the best fold model on the ADDM data results in almost every line in a record being labeled as ASD. Consequently, regardless of the threshold, records are consistently labeled as ASD, leading to very poor performance. The mixed data set shows the best performance compared to any training scheme with the individual datasets.

The performance of this mixed dataset is only slightly worse than that of our transparent approach when the highest threshold is used. As Fig. 4 shows, the performance of the transparent model is higher; only with the mixed dataset and the highest threshold does the black-box model come close to the transparent model.

**5. Discussion**

In this work, we focused on two main goals: evaluating the transfer of knowledge and comparing a transparent and a black-box approach. Related to this first goal, we compared the performance of a BioBERT model trained sequentially on two datasets or the mixed dataset. The mixed data model seems stabler across folds and datasets, and its F-value after training is slightly higher than that of the sequentially trained models. However, increases in performance are small and not significant for every metric. There are also differences depending on the order and type of datasets, indicating the need for further experimentation. For example, we employed different training and tuning learning rates, as is commonly done, with a higher learning rate and a reduced tuning rate. Additional evaluations with similar learning rates in sequences 1 and 2 might increase performance.

Our second goal centers on providing a transparent diagnostic process by focusing on intermediate steps. Our approach relies on assigning labels to intermediate steps in the ASD diagnostic process, creating a triple advantage. First, because the ASD diagnosis requires only one example of a behavior for a criterion, the model does not have to be perfect in labeling behaviors, and there is no single point of failure. Even when several behaviors are incorrectly labeled, the final decision can still be correct. Second, because a clinician can review these behaviors and associated labels, detecting incorrect labels is easy, as is subsequently adjusting the final diagnosis. Human review is possible and efficient. Finally, the black box model is more vulnerable to a common problem, i.e., the tendency to attach value to irrelevant information, even leading to spurious correlations. Sentences such as "*He is here for evaluation of possible autism*" or "*In special preschool since age 3*" were incorrectly labeled as ASD by the black-box model.

**Table 6**
Transparent model results at case level for best models for sequential and mixed data training.

|  | Sequence 1 | | | Sequence 2 | | | | | |
| --- | --- | --- | --- | --- | --- | --- | --- | --- | --- |
| Initial Training on | **ADDM** | | | **EHR** | | | **Mixed** | | |
|  | Accuracy | Sensitivity | Specificity | Accuracy | Sensitivity | Specificity | Accuracy | Sensitivity | Specificity |
| Test fold used for comparing models | **ADDM Fold 4** | | | **EHR Fold 9** | | | **Mixed Fold 7** | | |
| Best ADDM Model | 0.90 | 0.92 | 0.86 | 0.90 | 1 | 0.71 | 0.93 | 0.87 | 0.96 |
| Best EHR Modl | 0.93 | 0.67 | 1 | 0.97 | 0.92 | 0.98 | 0.95 | 0.97 | 0.94 |
| Best Mixed Model | 0.97 | 0.96 | 0.98 | 0.96 | 1 | 0.95 | 0.98 | 0.96 | 0.99 |
| Further Tuning on | **EHR** | | | **ADDM** | | | | | |
| Test fold used for comparing models |  | | | | | | | | |
| Best ADDM Model | 0.90 | 0.92 | 0.86 | 0.75 | 0.69 | 0.86 | | | |
| Best EHR Model | 0.98 | 0.92 | 1 | 0.97 | 0.83 | 1 | | | |

**Table 7**
Black-box model results at case level for sequential and mixed data training.

| | Sequence 1 | | | Sequence 2 | | | | | |
|---|---|---|---|---|---|---|---|---|---|
| Initial Training on | **ADDM** | | | **EHR** | | | **Mixed** | | |
| Threshold | Accuracy | Sensitivity | Specificity | Accuracy | Sensitivity | Specificity | Accuracy | Sensitivity | Specificity |
| 0.2 | 0.69 | 1 | 0.03 | 0.83 | 0.99 | 0.73 | 0.78 | 1 | 0.63 |
| 0.4 | 0.72 | 0.99 | 0.16 | 0.87 | 0.95 | 0.83 | 0.89 | 0.99 | 0.82 |
| 0.5 | 0.75 | 0.99 | 0.25 | 0.89 | 0.94 | 0.87 | 0.91 | 0.99 | 0.85 |
| 0.6 | 0.77 | 0.98 | 0.33 | 0.89 | 0.88 | 0.90 | 0.92 | 0.98 | 0.89 |
| 0.8 | 0.81 | 0.86 | 0.70 | 0.78 | 0.50 | 0.96 | 094 | 0.90 | 0.96 |
| Further Tuning on | **EHR** | | | **ADDM** | | | | | |
| 0.2 | 0.78 | 0.99 | 0.71 | 0.68 | 1 | 0 | | | |
| 0.4 | 0.82 | 0.96 | 0.77 | 0.68 | 1 | 0 | | | |
| 0.5 | 0.83 | 0.92 | 0.80 | 0.68 | 1 | 0 | | | |
| 0.6 | 0.85 | 0.84 | 0.86 | 0.68 | 1 | 0 | | | |
| 0.8 | 0.88 | 0.61 | 0.97 | 0.68 | 1 | 0 | | | |

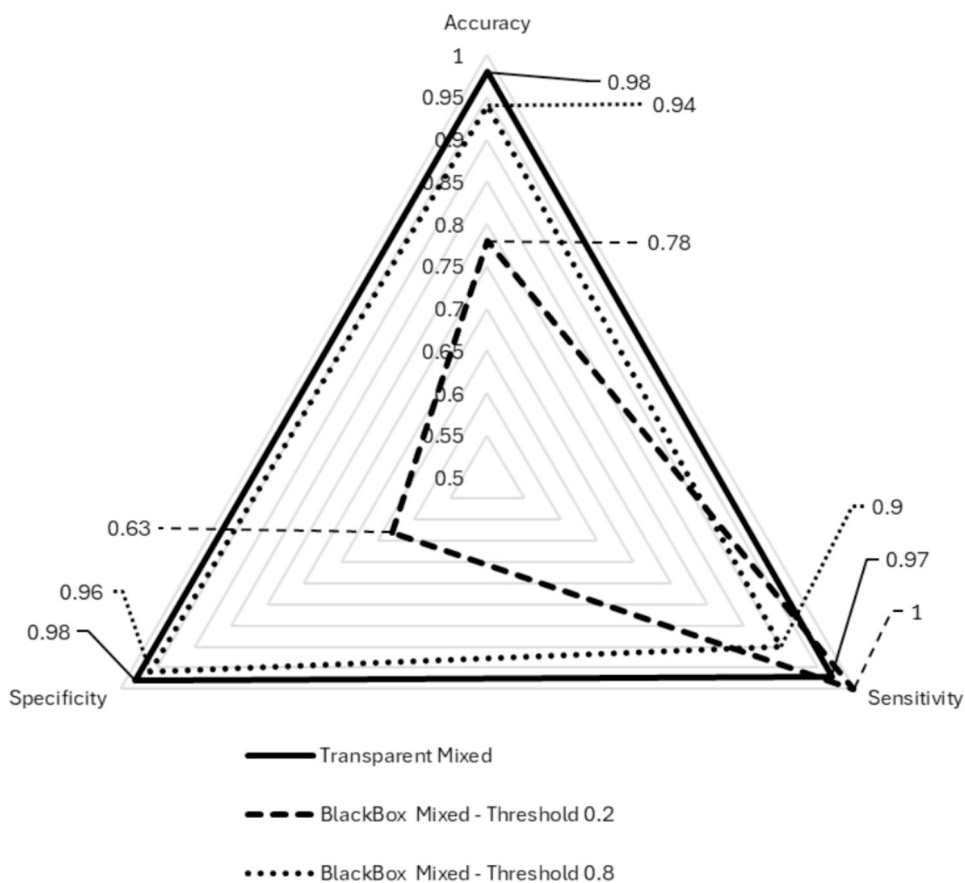

**Fig. 4.** Comparison of Transparent and Black-Box Models for Mixed Data Training.

Although the transparent model offers significant advantages, one main disadvantage is the time and effort needed to create the data. Using the black-box model requires only a case-level label, and no dataset needs to be created with line-level labels, making this a faster and cheaper approach.

This work also shows that training order matters. Transferring models to a new data set results in decreased performance. With additional training, the original model learns to perform well on the second dataset. Notably, sequence 2 showed very high performance on EHR data, achieving a score of 0.90 on the F1 measure, compared to 0.85 in Sequence 1. This may be due to the EHR data containing more negative examples (sentences without a label), resulting in very high precision. The change in learning rates may also have played a part, and more experimentation is needed. Our data using the best 'individual' models also shows how such models can show a significant change in performance with additional data. Ten-fold cross-validation is excellent for evaluating performance, but the final models are best trained on all data.

Our dataset was de-identified, and we believe that performance on a dataset containing unaltered text would be higher. With de-identification algorithms, even though the University of Arizona uses the gold standard, some information is incorrectly removed, while other information is riddled with extraneous nonsense. For example, in "*He is flexible and does xxx with changes in routine*" a crucial term is missing. In the line "*Last Name xxx 1\*\* is here today for a follow-up on her pinworms*", there are many irrelevant tokens. We did not apply spelling or grammar corrections, which might also improve results but would require significant additional resources.

It is worth pointing out that more examples do not always lead to better results. The criterion that was consistently labeled with the lowest performance (B3) was not the one for which the fewest examples were available. In fact, the fewest examples were available for B2, which is not the worst or second-worst performing criterion. We are currently experimenting with synthetic data to investigate whether dataset features and dataset size can provide insight [34].

Finally, compared to our earlier work using LSTM and GRU in various ensembles [46], there is less variance in performance per criterion, and overall performance is higher. In earlier work, the best F1 score was 0.58 for an inclusive-or combination of the two best ML algorithms. The precision was 61 % and the recall was 58 %, which is lower than any algorithm version in this study, showing the superiority of the BioBERT model.

## 6. Limitations

This work has several limitations related to the amount of training, models used, and analysis. First, we chose to use the best model of the ten models for further training and comparisons. We could also have retrained on ten folds and used this model for further tuning and comparisons. However, our intent in choosing one model was to capture a clear and strong signal and compare across stable test datasets that were not part of the training. For deployment purposes, rather than research, the final model would be retrained on all ten folds (not nine).

Second, better results may have been achieved with more training or by using more sophisticated models, e.g., by adding attention layers. Different parameters should also be tested and compared; for instance, varying learning rates, especially with the second dataset, may provide insights into optimal learning and adjusting versus overpowering or erasing the previous model.

Finally, we did not include an analysis of the style and content of the text itself. For example, we lack measures such as copy and paste detection, which is common in EHR, analysis of main topics, or frequencies of linguistic features. We intend to perform such an analysis in the future, as it may help explain the performance differences between criteria, since a simple example count is insufficient as an explanation.

## 7. Conclusion

In this work, we compared the transfer of a BioBERT model across different datasets. When adding new data, training on this new dataset is essential to avoid deteriorating performance. However, training on a mixed dataset leads to slightly better results. In practice, this may not always be possible. Further work is needed on using different training and tuning learning rates, as well as the effect of transferring models across more than two datasets.

We applied the BioBERT model to transparent and black-box approaches. The transparent approach performed better, as it focused on recognizing specifically relevant behaviors. In practice, this method is also more valuable, as it shows why a label was assigned to a case. However, a black-box approach requires much less effort to implement.

Finally, the BioBERT models demonstrated superior performance compared to older work using other deep learning models applied individually or as ensembles.

## Declaration of Generative AI

No generative AI was used in the writing of this paper or for the creation of the figures. Grammarly was used for editing.

## CRediT authorship contribution statement

**Gondy Leroy:** Supervision, Project administration, Investigation, Formal analysis, Validation, Resources, Methodology, Funding acquisition, Conceptualization, Writing – original draft, Writing – review & editing. **Prakash Bisht:** Conceptualization, Software. **Sai Madhuri Kandula:** Software. **Nell Maltman:** Resources, Supervision, Data curation. **Sydney Rice:** Resources.

## Declaration of competing interest

The authors have no competing interests to declare.

## Acknowledgements

This project was supported by grant number R01MH124935 from the National Institute of Mental Health. Part of the data presented were collected by the Centers for Disease Control (CDC) and Prevention Autism and Developmental Disabilities Monitoring (ADDM) Network supported by CDC Cooperative Agreement Number 5UR3/DD000680.

## Appendix A. DSM-5 Criterion Labels

**A Labels: Persistent deficits in social communication and social interaction across multiple contexts, as manifested by the following, currently or by history (examples are illustrative, not exhaustive, see text)**

A1. Deficits in social-emotional reciprocity, ranging, for example, from abnormal social approach and failure of normal back-and-forth conversation; to reduced sharing of interests, emotions, or affect; to failure to initiate or respond to social interactions

A2. Deficits in nonverbal communicative behaviors used for social interaction, ranging, for example, from poorly integrated verbal and nonverbal communication; to abnormalities in eye contact and body language or deficits in understanding and use of gestures; to a total lack of facial expressions and nonverbal communication

A3. Deficits in developing, maintaining, and understanding relationships, ranging, for example, from difficulties adjusting behavior to suit various social contexts; to difficulties in sharing imaginative play or in making friends; to absence of interest in peers

**B Labels: Restricted, repetitive patterns of behavior, interests, or activities, as manifested by at least two of the following, currently or by history (examples are illustrative, not exhaustive; see text):**

B1. Stereotyped or repetitive motor movements, use of objects, or speech (e.g., simple motor stereotypies, lining up toys or flipping objects, echolalia, idiosyncratic phrases).

B2. Insistence on sameness, inflexible adherence to routines, or ritualized patterns or verbal nonverbal behavior (e.g., extreme distress at small changes, difficulties with transitions, rigid thinking patterns, greeting rituals, need to take same route or eat food every day).

B3. Highly restricted, fixated interests that are abnormal in intensity or focus (e.g, strong attachment to or preoccupation with unusual objects, excessively circumscribed or perseverative interest).

B4. Hyper- or hyporeactivity to sensory input or unusual interests in sensory aspects of the environment (e.g., apparent indifference to pain/temperature, adverse response to specific sounds or textures, excessive smelling or touching of objects, visual fascination with lights or movement).

## Data availability

The data used to train the ML algorithms and to evaluate the diagnostic tests are not publicly available due to the ADDM Network Data Confidentiality and Security Agreement and the EHRs limitation of data to university researchers. The parameters used to train the ML algorithm are described in the manuscript.